\documentclass[10pt,twocolumn,letterpaper]{article}

\usepackage{wacv}              

\usepackage{array}
\usepackage{graphicx}
\usepackage{amsmath}
\usepackage{amssymb}
\usepackage{booktabs}
\usepackage{colortbl} 
\usepackage{xcolor}
\usepackage{multirow}
\usepackage{makecell}

\usepackage[pagebackref,breaklinks,colorlinks]{hyperref}

\usepackage[capitalize]{cleveref}
\crefname{section}{Sec.}{Secs.}
\Crefname{section}{Section}{Sections}
\Crefname{table}{Table}{Tables}
\crefname{table}{Tab.}{Tabs.}

\def\wacvPaperID{*****} 
\def\confName{WACV}
\def\confYear{2025}

\graphicspath {{figures/}}
\begin{document}

\title{NCAP: Scene Text Image Super-Resolution with Non-CAtegorical Prior}

\author{Dongwoo Park and SUK PIL KO\\
THINKWARE Corporation, Republic of Korea\\
{\tt\small \{infinity7428, spko\}@thinkware.co.kr}
}
\maketitle


\begin{abstract}
Scene text image super-resolution (STISR) enhances the resolution and quality of low-resolution images. Unlike previous studies that treated scene text images as natural images, recent methods using a text prior (TP), extracted from a pre-trained text recognizer, have shown strong performance. However, two major issues emerge: (1) Explicit categorical priors, like TP, can negatively impact STISR if incorrect. We reveal that these explicit priors are unstable and propose replacing them with Non-CAtegorical Prior (NCAP) using penultimate layer representations. (2) Pre-trained recognizers used to generate TP struggle with low-resolution images. To address this, most studies jointly train the recognizer with the STISR network to bridge the domain gap between low- and high-resolution images, but this can cause an overconfidence phenomenon in the prior modality. We highlight this issue and propose a method to mitigate it by mixing hard and soft labels.
Experiments on the TextZoom dataset demonstrate an improvement by 3.5\%, while our method significantly enhances generalization performance by 14.8\% across four text recognition datasets.
Our method generalizes to all TP-guided STISR networks.
\end{abstract}

\begin{figure*}[t]
	\centerline{\includegraphics[width=0.9\textwidth]{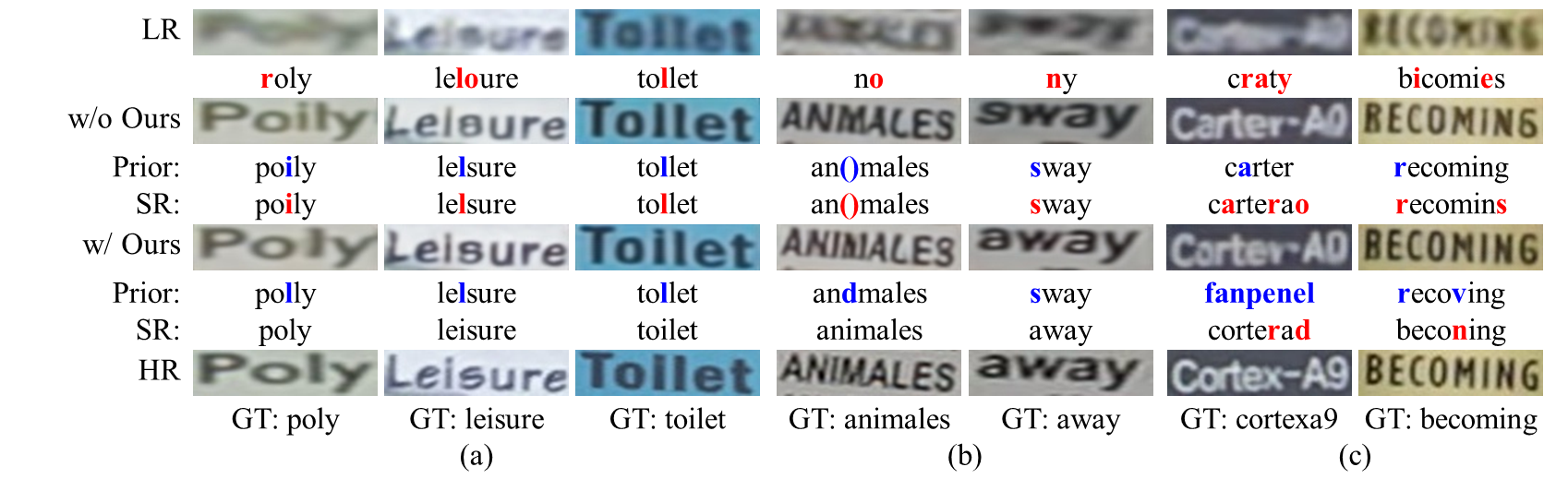}}
	\caption{Examples illustrating the negative impact of prior knowledge on an STISR task.
(a), (b), and (c) of w/ and w/o Ours refer to the results of using CRNN \cite{shi2016end} as a prior generator (pre-trained text recognizer) in TATT \cite{ma2022text}, PARSeq \cite{bautista2022scene} as a prior generator in TATT \cite{ma2022text}, and LEMMA \cite{guo2023towards}, respectively.
Even with various STISR networks and a prior generator, the wrong guidance of the explicit prior still appears. \textcolor{blue}{Blue} indicates the characters that can be influenced by prior knowledge in the STISR results. \textcolor{red}{Red} indicates wrong recognition results. Without our method, prior knowledge negatively influences the STISR results; however, with our proposed method, this negative influence can be effectively eliminated. Prior refers to the argmaxed text prior and SR refers to the recognition result of the SR image.}
\end{figure*}

\section{Introduction}
Recognizing and interpreting the text contained in images is a crucial task.
Despite numerous advancements in areas such as optical character recognition (OCR), scene text detection (STD), and scene text recognition (STR), the acquisition of low-resolution images is inevitable due to various factors, including the quality of the lens, motion blur, low light, occlusion, noise, etc.
Thus, recognizing text in such images remains a challenging task.
To reliably recover missing structural details from such images, scene text image super-resolution (STISR) is employed as a pre-processing step to address low-resolution issues.
Many studies on scene text images achieve high performance by treating the text prior as an additional modality to leverage text semantics rather than interpreting it as natural scene images.
However, due to the explicit nature of the text prior, misclassified categories within the prior knowledge have the potential to harm the STISR task.
Thus, some studies address the challenges associated with the text prior.
For instance, TPGSR \cite{ma2023text} is a pioneering work that suggests incorporating feedback from a text recognizer, referred to as the text prior, as an additional modality.
They also acknowledge accuracy issues with the text prior. To address this, they attempted to complement it by employing the teacher-student structure to distill the text logits from the teacher recognizer.
DPMN \cite{zhu2023improving} introduced a method to complement a pre-trained STISR network using two explicit image-level priors, namely the text mask and graphical recognition results.
C3-STISR \cite{zhao2022c3} improves the prior knowledge using the text prior and its transformed visual cue and linguistic information.
LEMMA \cite{guo2023towards} enhances recognizer accuracy by minimizing errors in ground-truth text labels through fine-tuning loss.

However, the aforementioned methods focus solely on improving the performance of the text prior without addressing the fundamental instability inherent in the text prior.
As a pioneering work, we propose Non-CAtegorical Prior (NCAP)—penultimate layer representations processed by additional adapters—instead of relying on the text prior from a pre-trained text recognizer.
Figure 1 illustrates the negative impact of incorrect prior knowledge on the STISR result.
Our proposed method effectively addresses the aforementioned problem by transforming explicit prior knowledge into an implicit form.

Another issue arises during joint training of the STISR network with a recognizer to minimize the domain gap between low-resolution (LR) and high-resolution (HR) images using ground-truth labels, which results in the overconfidence issue.
The performance of existing pre-trained text recognizers degrades when applied to LR images.
When training the recognizer along with the STISR network to overcome the domain gap between LR and HR images, training the recognizer with soft labels can lead to inaccurate prior knowledge, while training with hard labels leads to an overconfidence issue.
Hence, we compare the differences between the hard labels and soft labels used in the existing loss function.
We provide both theoretical and empirical analyses of the loss function.
In conclusion, we propose a loss function that mitigates the overconfidence issue by employing a linear combination of the hard label and the soft label in a simple yet effective way.

The contributions of our proposed method are as follows:
(1) We present NCAP, a more information-rich and stable knowledge, achieved by replacing the existing unstable text logits with penultimate layer representations. Our proposed NCAP, including its adapters, requires only $0.3\%$ additional parameters. Compared with existing methods, NCAP fundamentally resolves explicit information issues.
(2) Our work identifies the overconfidence issue that arises from joint training with ground-truth hard labels. We address this issue through a simple but effective approach employing a linear combination of softened Kullback-Leibler (KL) divergence and cross-entropy losses.

\begin{figure*}[t]
  \centerline{\includegraphics[width=0.9\linewidth]{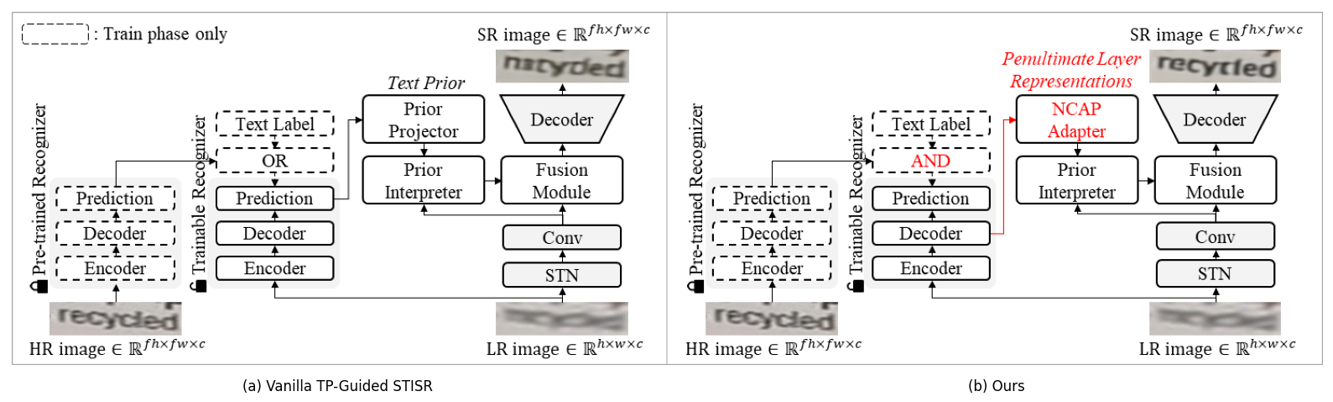}}
  \caption{Overall architecture. We enhance the previous TP-guided STISR network by introducing a loss function that incorporates linear combinations of hard labels and soft labels, along with NCAP, which utilizes penultimate layer representations as prior knowledge.}
\end{figure*}

\section{Related Works}
\subsection{Scene Text Recognition (STR)}
Scene text recognition aims to recognize character sequences from scene text images.
Convolutional Recurrent Neural Network (CRNN) \cite{shi2016end} predicts character sequences using Connectionist Temporal Classification (CTC) \cite{graves2006connectionist} loss through a combination of Convolutional Neural Networks (CNNs) and Recurrent Neural Networks (RNNs).
ASTER \cite{shi2018aster} is a pioneering study that introduces Thin-Plate Spline (TPS) transformation for handling spatially transformed text.
MORAN \cite{luo2019moran} proposed a solution for irregular text by combining multi-object rectification and attention-based sequence prediction.
ABINet \cite{fang2021read} corrects predictions through external language modeling.
PARSeq \cite{bautista2022scene} does not use language models separately, and performs permuted sequence modeling while removing the dependence of sequence order through iteratively permuted internal language modeling.
Using text as prior knowledge through STR is beneficial for STISR, but it remains a double-edged sword.
The problem lies in the negative impact of the explicit categorical information from the recognizer, referred to as the text prior (TP), which can lead to incorrect STISR results when influenced by an inaccurate TP.
To address this issue, we propose NCAP, which uses implicit prior knowledge.

\subsection{Scene Text Image Super-Resolution (STISR)}
While scene image super-resolution (SISR) primarily emphasizes enhancing the visual quality of images, STISR has recently shifted its focus to not only improving visual quality but also prioritizing the ultimate goal of enhancing text readability.
TSRN \cite{wang2020scene} not only introduces the most commonly used dataset for STISR but also captures the sequential information of the text through consecutive CNN-BiLSTM layers.
Subsequently, TBSRN \cite{chen2021scene} incorporates position- and content-aware losses with an attention map.
TPGSR \cite{ma2023text} introduces a text prior, which represents text categorical information obtained from a pre-trained text recognizer.
TATT \cite{ma2022text} integrates a global attention mechanism to address the imperfections caused by spatially deformed images, mitigating the structural limitations of CNNs.
DPMN \cite{zhu2023improving} highlights the significance of global structure by leveraging two visual priors within a fixed pre-trained STISR network.
LEMMA \cite{guo2023towards} introduces a novel approach that specifically exploits character location information.

Despite the advantage of being able to read text by employing a pre-trained recognizer, the aforementioned method still has issues.  
As a result, TP methods that use fixed pre-trained recognizers struggle to overcome the domain gap with low-resolution images.
Even when the recognizer is jointly trained with the STISR network, existing methods suffer from accuracy problems or overconfidence issues. Therefore, we propose a loss that mixes hard and soft labels to effectively resolve these issues.

\section{Methodology}
\subsection{Overview}
Figure 2 illustrates the structure of our proposed method, which contains two main different components compared to the existing method.
One is NCAP and its adapters, which replace the unstable text prior with stable penultimate layer representations. The other is the mixing of hard and soft labels operation, which is intended to solve the overconfidence issue.
The training input consists of a low-resolution (LR) image $I_{LR}$, a high-resolution (HR) image $I_{HR}$, and a ground-truth text label $y_{gt}$. The objective is to enhance the LR image $I_{LR}\in \mathbb{R}^{h\times w\times 3}$ to a super-resolution (SR) image $I_{SR}\in \mathbb{R}^{fh\times fw\times 3}$ of the same size as the HR image $I_{HR}\in \mathbb{R}^{fh\times fw\times 3}$.
The LR image is input to a Shallow CNN for feature extraction and simultaneously to the student recognizer for NCAP.
This generates the extracted image features $f_I \in \mathbb{R}^{h \times w \times c}$ and the penultimate layer representations $h$ of the recognizer.
The generated feature $h$ is projected again as a $f_{NCAP}$ by the adapters. After the fusion operation of the image feature $f_{I}$ and NCAP $f_{NCAP}$, the combined features are input to the decoder module to generate a restored image.

\subsection{Non-CAtegorical Prior (NCAP)}
The existing studies have demonstrated the pivotal role of strong prior knowledge.
TPGSR \cite{ma2023text} demonstrated better performance than TSRN \cite{wang2020scene} without utilizing prior knowledge, while also introducing the use of prior knowledge as an effective approach. This concept has since been utilized in other methods, such as TATT \cite{ma2022text} and C3-STISR \cite{zhao2022c3}, achieving remarkable results.

However, all of these approaches failed to address the fundamental problem: inaccuracy.
This can lead to incorrect guidance in the STISR task due to category inconsistency in the text prior.
To address this issue, we propose a method of using penultimate layer representations, which are in the stage before being converted to text probability information, as prior knowledge. This information is not only category-free but also contains rich information and can be processed through additional adapters with only about $0.3\%$ overhead.
In existing works that use the text prior, hidden representation vectors $h\in \mathbb{R}^{L \times embed}$ are generated by predicting the sequence of characters using the pre-trained text recognizer.
Then, it goes through the process of mapping $h$ to the text logits through a prediction layer $W \in \mathbb{R}^{embed \times |A|}$.
$embed$ denotes the dimension of the penultimate layer representations.
$L$ is the length of the pre-specified maximum character composed of categorical probability vectors with size $|A|$, and $A$ denotes the alphabet set, which is composed of alphanumeric characters, and in the case of PARSeq \cite{bautista2022scene}, it also includes punctuation marks.
Finally, the text prior $f_{TP}$ is completed through the projection layer $W_{proj} \in \mathbb{R}^{|A| \times C}$.
However, we directly use the category-free hidden representation vectors $h$ that have not been converted to a probability distribution by projecting them through additional adapters $W_{adapters}$. $C$ denotes the dimension of the prior feature.

\begin{equation}
f_{NCAP} = \mathrm{PReLU}(\mathrm{PReLU}( h \cdot W_{adapter_{1}}) \cdot W_{adapter_{2}})
\end{equation}
Equation 1 illustrates the process of generating NCAP from penultimate layer representations. $h \in \mathbb{R}^{L \times embed}$ is mapped to $\mathbb{R}^{L \times \frac{embed}{2}}$ by $W_{adapter_{1}}$ and then mapped to $\mathbb{R}^{L \times C}$ by $W_{adapter_{2}}$.
Parametric Rectified Linear Unit ($\mathrm{PReLU}$) \cite{he2015delving} is an activation function that provides learnable gradients in the negative range.

\begin{figure*}[h!]
	\centerline{\includegraphics[width=0.8\linewidth]{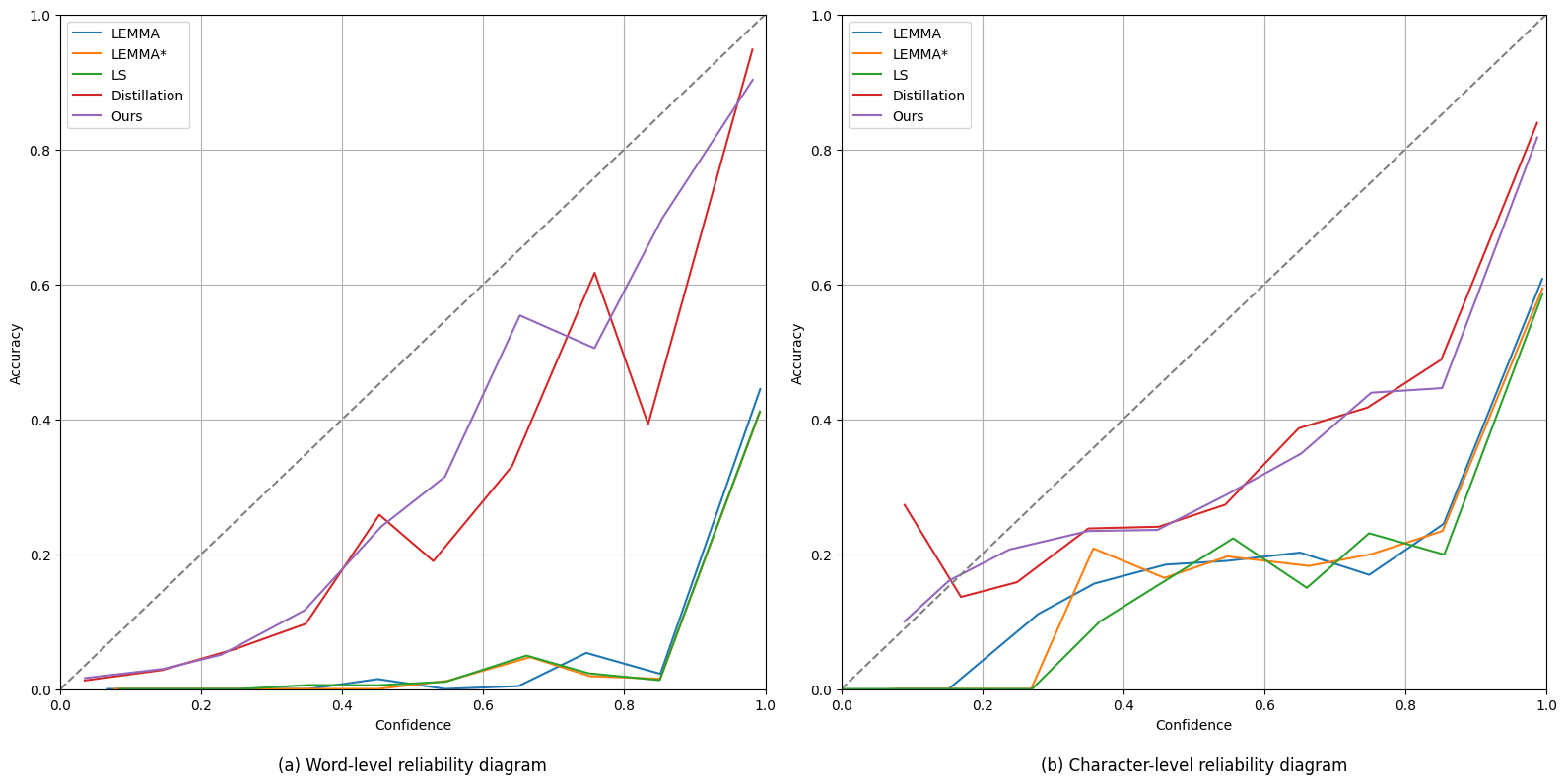}}
	\caption{Word- and character-level reliability diagram. LEMMA \cite{guo2023towards} is the result visualized with the pre-trained weights of the official code, LEMMA$^{*}$ corresponds to the result of re-training the model using the official code of LEMMA \cite{guo2023towards}, LS \cite{he2015delving} represents a label smoothing technique, and Distillation represents the result of training a model using a loss function that eliminates the learning process with hard labels, opting to learn from soft labels instead. Ours is the result of a model trained with a linear combination loss of hard and soft labels. Please refer to the supplementary materials for the calculation of character-level reliability.}
	\vspace{-0.2cm}
\end{figure*}

\subsection{Overcoming the Domain Gap between LR-HR Images While Avoiding Overconfidence}
As TPGSR \cite{ma2023text} revealed the low performance of a fixed recognizer, the method of jointly training the text recognizer with the STISR network to overcome the domain gap between LR and HR images became popular.
Therefore, to reduce the domain gap and improve performance, it is inevitable to train recognizers with the STISR network.
As joint training with the STISR network improves performance, it also introduces new challenges, such as the overconfidence issue. This problem is clearly illustrated in Figure 3, where fine-tuning the recognizer with ground-truth hard labels leads to overconfidence in the prior modality.
Depending on the ground-truth label used, methods can be broadly classified into two categories. One approach is to train a student recognizer with soft labels generated from the teacher recognizer through a teacher-student structure, as seen in TPGSR \cite{ma2023text}, TATT \cite{ma2022text}, and C3-STISR \cite{zhao2022c3}. Another approach is to train directly using ground-truth hard labels, exemplified by LEMMA \cite{guo2023towards}.
However, both methods also have their own problems.
In the method that relies only on soft labels, there is an increase in inconsistency in the prior knowledge, while in the method relying only on hard labels, overconfidence is induced, ultimately leading to a decrease in performance.
To address the overconfidence issue and enhance performance, we propose a method that leverages the strengths of both approaches by linearly combining hard and soft labels.

\begin{figure*}[t]
	\centerline{\includegraphics[width=0.9\linewidth]{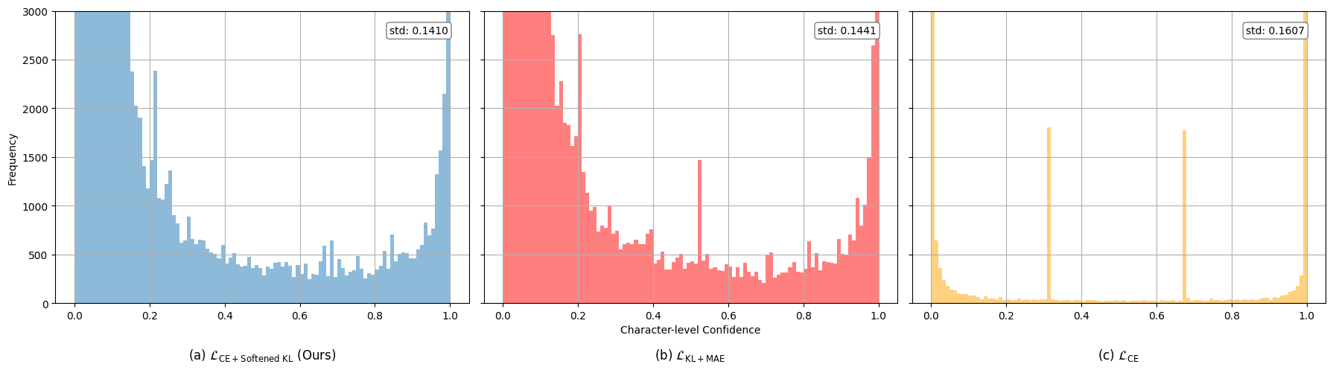}}
	\caption{Results of character-level distribution difference of the text logits for training data by each loss function. (a) is a linear combination of softened KL divergence loss and cross-entropy loss, (b) involves KL divergence loss along with MAE loss, which are used in TPGSR \cite{ma2023text} and TATT \cite{ma2022text}, and (c) corresponds to the cross-entropy loss used in LEMMA \cite{guo2023towards}.}
	\vspace{-0.2cm}
\end{figure*}

 \textbf{Theoretical support.}
Kullback-Leibler (KL) divergence loss combined with mean absolute error (MAE) loss has been utilized in both TPGSR \cite{ma2023text} and TATT \cite{ma2022text}. To analyze its impact, we compare KL divergence loss alone with KL divergence loss combined with MAE loss by examining the gradients generated through the partial derivatives of these loss functions.
Additionally, our goal is to explore the conditions and penalties that lead to effective performance improvements while addressing the issue of overconfidence.

 \textbf{Kullback-Leibler Divergence Loss.}
\begin{equation}
 \frac{\partial\mathcal{L}^{s}_{KL}}{\partial z_{i}}=p^{s}_{i}-p^{t}_{i}
\end{equation}
The gradient of $\mathcal{L}^{s}_{KL}$ with respect to a student's logit value $z_{i}$ is defined as in Equation (2), and since there is no case for $i=true$, so all indices $i$ have the same gradient range. Let $p^{s}_{i}$ be the student's prediction and $p^{t}_{i}$ the teacher's prediction at the $i$-th class.

\textbf{Kullback-Leibler Divergence Loss with Mean Absolute Error Loss.}
\begin{equation}
 \frac{\partial\mathcal{L}^{s}_{KL+MAE}}{\partial z_{i}}=p^{s}_{i}-p^{t}_{i}+\begin{cases}
    +1, & \text{if $z_{i}>t_{i}$}\\
    -1, & \text{if $z_{i}<t_{i}$}\\
  \end{cases}
\end{equation}
$for \ z_i>t_i,$
\begin{equation}
\partial\mathcal{L}^{s}_{KL+MAE} = p^{s}_{i}-(-1+p^{t}_{i})
\end{equation}
$for \ z_i<t_i,$
\begin{equation}
\partial\mathcal{L}^{s}_{KL+MAE} = p^{s}_{i}-(1+p^{t}_{i})
\end{equation}

$t_{i}$ represents the teacher's logit value.
Equations (3) to (5) define the gradient of $\mathcal{L}^{s}_{KL+MAE}$ with respect to a student's logit value $z_{i}$, where both MAE loss and KL divergence loss, as used in TPGSR~\cite{ma2023text} and TATT~\cite{ma2022text}, are applied. The inclusion of MAE loss imposes, on average, larger gradients than using only the KL divergence loss from Equation (2).
This can sharpen soft label distributions. We found that sharpening inaccurate soft labels destabilizes performance.

\begin{table}[t]
    \centering\scriptsize
    \resizebox{\linewidth}{!}{
        \begin{tabular}{l|ccc|c}
            \toprule
            \textbf{Loss Function} & \textbf{ASTER \cite{shi2018aster}} & \textbf{MORAN \cite{luo2019moran}} & \textbf{CRNN \cite{shi2016end}} & \textbf{Average} \\
            \midrule
            w/o loss & 59.9\% & 56.3\% & 48.4\% & 54.9\% \\
            $\mathcal{L}_{KL+MAE}$ & 66.6\% & 64.1\% & 57.1\% & 62.6\% \\
            $\mathcal{L}_{CE}$ & 66.0\% & 63.2\% & 56.3\% & 61.8\% \\
            $\mathcal{L}_{CE\ with\ LS}$ & 66.5\% & 62.8\% & 55.6\% & 61.6\% \\
            $\mathcal{L}_{CE+KL}$ & 66.1\% & 64.5\% & 56.8\% & 62.5\% \\
            $\mathcal{L}_{CE+KL+MAE}$ & \textbf{66.9\%} & 63.9\% & \textbf{57.7\%} & 62.8\% \\
            $\mathcal{L}_{CE+Softened\ KL}$ & 66.8\% & \textbf{64.6\%} & 57.5\% & \textbf{63.0\%} \\
            \bottomrule
        \end{tabular}
    }
    \caption{Recognition accuracies of the TextZoom \cite{wang2020scene} for each loss function using ASTER \cite{shi2018aster}, MORAN \cite{luo2019moran}, and CRNN \cite{shi2016end}. w/o loss trains the network using only the loss of the STISR network without any loss function for the recognizer. $\mathcal {L}_{KL+MAE}$ is the loss function proposed in TPGSR \cite{ma2023text}, and $\mathcal {L}_{CE}$ is the loss function used in LEMMA \cite{guo2023towards}. LS represents a label smoothing \cite{he2015delving} technique. Average refers to average accuracy.}
    \label{tab:loss_functions}
    \vspace{-0.2cm}
\end{table}

Figure 4 shows the character-level distribution plot on the TextZoom \cite{wang2020scene} train data for each loss. Our proposed method has the lowest standard deviation of 0.1410 and shows a smooth distribution.
Table 1 displays the results when each loss function is applied to the STISR network. 
The best performance is achieved when both ground-truth hard labels and the teacher's soft labels are utilized together, particularly when the soft labels exhibit a smoother profile.

We made three key observations:
(1) Despite the overconfidence induced by the ground-truth hard label, it remains an essential factor for performance enhancement.
(2) Soft labels effectively mitigate overconfidence.
(3) Mixing hard and soft labels shows greater effectiveness. The performance improves as the distribution of soft labels becomes smoother.
Thus, we propose a linear combination of softened KL divergence loss by temperature scaling and cross-entropy loss as a new loss function.

\begin{table*}[ht]
    \centering\scriptsize
    \begin{tabular}{l|cc|cc|cc}
        \toprule
        \multirow{2}{*}{\textbf{Method}} & \multicolumn{2}{c|}{\textbf{Prior Error Rate}} & \multicolumn{2}{c|}{\textbf{SR Error Rate}} & \multicolumn{2}{c}{\textbf{Pearson Correlation}} \\
        \cmidrule(lr){2-3} \cmidrule(lr){4-5} \cmidrule(lr){6-7}
        & \textbf{WER} & \textbf{CER} & \textbf{WER} & \textbf{CER} & \textbf{WER} & \textbf{CER} \\
        \midrule
        TATT \cite{ma2022text} & 52.3\% & 32.2\% & 47.2\% & 30.7\% & 0.7146 & 0.8026 \\
        TATT \cite{ma2022text} w/ Ours & \textbf{37.4\%} & \textbf{21.3\%} & \textbf{43.3\%} & \textbf{27.1\%} & \textbf{0.6626} & \textbf{0.7359} \\
        \multicolumn{1}{c|}{$\Delta$} & \multicolumn{1}{c}{\color{blue} \textbf{-14.9\%}} & \multicolumn{1}{c|}{\color{blue} \textbf{-11.0\%}} & \multicolumn{1}{c}{\color{blue} \textbf{-3.9\%}} & \multicolumn{1}{c|}{\color{blue} \textbf{-3.6\%}} & \multicolumn{1}{c}{\color{blue} \textbf{-7.3\%}} & \multicolumn{1}{c}{\color{blue} \textbf{-8.3\%}} \\
        \midrule
        LEMMA \cite{guo2023towards} & 76.1\% & 58.3\% & 44.0\% & 28.3\% & 0.3465 & 0.4580 \\
        LEMMA \cite{guo2023towards} w/ Ours & \textbf{77.6\%} & \textbf{60.5\%} & \textbf{42.1\%} & \textbf{26.9\%} & \textbf{0.3279} & \textbf{0.3052} \\
        \multicolumn{1}{c|}{$\Delta$} & \multicolumn{1}{c}{\color{red} \textbf{+1.5\%}} & \multicolumn{1}{c|}{\color{red} \textbf{+2.2\%}} & \multicolumn{1}{c}{\color{blue} \textbf{-2.0\%}} & \multicolumn{1}{c|}{\color{blue} \textbf{-1.3\%}} & \multicolumn{1}{c}{\color{blue} \textbf{-5.4\%}} & \multicolumn{1}{c}{\color{blue} \textbf{-33.4\%}} \\
        \bottomrule
    \end{tabular}
    \caption{Comparison of Pearson correlation coefficients between the prior text logits error rate and restored image text recognition error rate. WER and CER represent word error rate and character error rate, respectively.}
    \label{tab:pearson_correlation}
    \vspace{-0.2cm}
\end{table*}

\begin{table}[ht]
    \centering\scriptsize
    \begin{tabular}{l|cc|cc}
        \toprule
        \textbf{Method} & \textbf{NCAP} & \textbf{\begin{tabular}[c]{@{}c@{}}Adapters\end{tabular}} & \textbf{MACs} & \textbf{\#Params} \\
        \midrule
        TATT \cite{ma2022text} &  &  & 4.60 G & 31.44 M \\
        \multirow{2}{*}{\begin{tabular}[c]{@{}l@{}}TATT  \cite{ma2022text} w/ Ours\end{tabular}} 
        & $\checkmark$ &  & 4.64 G & 31.52 M \\
        & $\checkmark$ & $\checkmark$ & 4.43 G & 31.52 M \\
        \multicolumn{1}{c|}{$\Delta$} & \multicolumn{2}{c|}{} & {\color{blue} \textbf{-3.7\%}} & {\color{red} \textbf{+0.3\%}} \\
        \midrule
        LEMMA \cite{guo2023towards} &  &  & 6.69 G & 39.75 M \\
        \multirow{2}{*}{\begin{tabular}[c]{@{}l@{}}LEMMA  \cite{guo2023towards} w/ Ours\end{tabular}} 
        & $\checkmark$ &  & 6.69 G & 39.90 M \\
        & $\checkmark$ & $\checkmark$ & 6.71 G & 39.90 M \\
        \multicolumn{1}{c|}{$\Delta$} & \multicolumn{2}{c|}{} & {\color{red} \textbf{+0.3\%}} & {\color{red} \textbf{+0.4\%}} \\
        \bottomrule
    \end{tabular}
    \caption{Results of computational complexity and increase in trainable parameters.}
    \label{tab:complexity_parameters}
    \vspace{-0.2cm}
\end{table}

\subsection{Training Objective}
We incorporate the proposed method into both TATT \cite{ma2022text}, and LEMMA \cite{guo2023towards}.
We do not alter the implementation details of the various loss functions for each method, but only modify the loss function, as shown in Equations (6) to (8), to address the overconfidence issue by linearly combining hard labels and soft labels.
$\alpha$ is a hyperparameter for balance.
Please refer to the supplementary materials for details on the various loss functions used in each method.
\begin{equation}
\mathcal{L}=(1-\alpha)\mathcal{L}_{CE}(p^{s},y)+\alpha \mathcal{L}_{KL}(p^{s}(\tau),p^{t}(\tau))
\end{equation}
\begin{equation}
\mathcal{L}_{CE}(p^{s},y)=-\sum_{i}{y_{i}\log{p^{s}_{i}}}
\end{equation}
\begin{equation}
\mathcal{L}_{KL}(p^{s}(\tau),p^{t}(\tau))=\beta \cdot \tau^{2} \sum_{i}{p^{t}_{i}(\tau) \log{\frac{p^{t}_{i}(\tau)}{p^{s}_{i}(\tau)}}}
\end{equation}
where $s$ indicates the student text recognizer, $t$ indicates the teacher text recognizer, $y$ is a ground-truth label, and $\tau$ is a temperature scaling parameter used for a smoother distribution.

\section{Experiments}
We introduce the experimental dataset, evaluation methods, and implementation details. We demonstrate the superiority of our proposed method through comparisons with state-of-the-art methods and various experiments.

\subsection{Datasets}
\textbf{Scene Text Image Super-Resolution Dataset.}
TextZoom \cite{wang2020scene} is a widely utilized dataset in STISR tasks. This dataset extracts the text-containing regions from two distinct super-resolution datasets, RealSR \cite{cai2019toward} and SR-RAW \cite{zhang2019zoom}, to create pairs of LR and HR images. The training set comprises 17,367 LR-HR image pairs with corresponding ground-truth text labels. The test set includes a total of 4,373 image pairs, distributed across three categories (1,619 for easy, 1,411 for medium, and 1,343 for hard) based on focal length. All LR images are resized to $16 \times 64$, and the HR images are adjusted to $32 \times 128$.

\textbf{Scene Text Recognition Dataset.}
To assess the robustness of our proposed method, we evaluate it on scene text recognition datasets: IIIT5K \cite{mishra2012scene}, ICDAR2015 \cite{karatzas2015icdar}, SVT \cite{wang2011end}, and SVTP \cite{phan2013recognizing}.
We assess the extent to which degraded images can be restored. These images are converted into LR images through manual degradation, following the approach used in previous studies such as TATT \cite{ma2022text} and LEMMA \cite{guo2023towards}. Additionally, we examine the generalization performance of models trained with the TextZoom \cite{wang2020scene}.

\subsection{Evaluation Metrics}
Following TPGSR \cite{ma2023text}, TATT \cite{ma2022text}, and LEMMA \cite{guo2023towards}, the performance of restored images is evaluated using three commonly used methods: text recognition accuracy, Peak Signal-to-Noise Ratio (PSNR), and Structure-Similarity Index Measure (SSIM).
In the case of accuracy, it is evaluated using pre-trained recognizers CRNN \cite{shi2016end},  ASTER \cite{shi2018aster}, and MORAN \cite{luo2019moran} as an evaluation metrics to evaluate readability, which is the ultimate goal of the STISR task.
PSNR and SSIM are used to evaluate the visual quality of the restored image.

\begin{table*}[ht]
    \centering\scriptsize
        \begin{tabular}{lccc|ccc|c}
            \toprule
            \textbf{Method} & \textbf{Loss} & \textbf{NCAP} & \textbf{Adapters} & \textbf{ASTER \cite{shi2018aster}} & \textbf{MORAN \cite{luo2019moran}} & \textbf{CRNN\cite{shi2016end}} & \textbf{Average} \\
            \midrule
            TATT \cite{ma2022text}           &           &           &           & 63.7\% & 59.4\% & 52.8\% & 58.6\% \\
            \multirow{3}{*}{TATT \cite{ma2022text} w/ Ours}    & \checkmark &           &           & 66.8\% & 63.3\% & 56.0\% & 62.0\% \\
                                             & \checkmark & \checkmark &           & 67.0\% & 63.8\% & 56.8\% & 62.5\% \\
                                             & \checkmark & \checkmark & \checkmark & \textbf{68.1\%} & \textbf{64.6\%} & \textbf{58.3\%} & \textbf{63.7\%} \\
            \midrule
            LEMMA \cite{guo2023towards}      &           &           &           & 66.0\% & 63.2\% & 56.3\% & 61.8\% \\
            \multirow{3}{*}{LEMMA \cite{guo2023towards} w/ Ours}   & \checkmark &           &           & 66.8\% & 64.6\% & 57.5\% & 63.0\% \\
                                             & \checkmark & \checkmark &           & 66.9\% & \textbf{65.1\%} & 58.0\% & 63.3\% \\
                                             & \checkmark & \checkmark & \checkmark & \textbf{67.9\%} & 65.0\% & \textbf{58.1\%} & \textbf{63.7\%} \\
            \bottomrule
        \end{tabular}
        \caption{Effectiveness of each module. Loss signifies the linear combination of the softened KL divergence and cross-entropy losses.}
        \label{tab:module_effectiveness}
        \vspace{-0.2cm}
\end{table*}

\begin{table*}[ht]
    \centering\scriptsize
        \begin{tabular}{l|cccc|cccc|cccc}
            \toprule
            \multirow{2}{*}{\makecell[c]{\textbf{Method}}} & \multicolumn{4}{c|}{\textbf{ASTER \cite{shi2018aster}}} & \multicolumn{4}{c|}{\textbf{MORAN \cite{luo2019moran}}} & \multicolumn{4}{c}{\textbf{CRNN \cite{shi2016end}}} \\
            \cmidrule(lr){2-5} \cmidrule(lr){6-9} \cmidrule(lr){10-13}
            & \textbf{Easy} & \textbf{Medium} & \textbf{Hard} & \textbf{Overall} & \textbf{Easy} & \textbf{Medium} & \textbf{Hard} & \textbf{Overall} & \textbf{Easy} & \textbf{Medium} & \textbf{Hard} & \textbf{Overall} \\
            \midrule
            Bicubic & 64.7\% & 42.4\% & 31.2\% & 47.2\% & 60.6\% & 37.9\% & 30.8\% & 44.1\% & 36.4\% & 21.1\% & 21.1\% & 26.8\% \\
	    \midrule
            TBSRN \cite{chen2021scene} & 75.7\% & 59.9\% & 41.6\% & 60.0\% & 74.1\% & 57.0\% & 40.8\% & 58.4\% & 59.6\% & 47.1\% & 35.3\% & 48.1\% \\
            TG \cite{chen2022text} & 77.9\% & 60.2\% & 42.4\% & 61.3\% & 75.8\% & 57.8\% & 41.4\% & 59.4\% & 61.2\% & 47.6\% & 35.5\% & 48.9\% \\
            TPGSR \cite{ma2023text} & 77.0\% & 60.9\% & 42.4\% & 60.9\% & 72.2\% & 57.8\% & 41.3\% & 57.8\% & 61.0\% & 49.9\% & 36.7\% & 49.8\% \\
            TPGSR-3 \cite{ma2023text} & 78.9\% & 62.7\% & 44.5\% & 62.8\% & 74.9\% & 60.5\% & 44.1\% & 60.5\% & 63.1\% & 52.0\% & 38.6\% & 51.8\% \\
            DPMN (+TATT) \cite{zhu2023improving} & 79.3\% & 64.1\% & 45.2\% & 63.9\% & 73.3\% & 61.5\% & 43.9\% & 60.4\% & 64.4\% & 54.2\% & 39.2\% & 53.4\% \\
            C3-STISR \cite{zhao2022c3} & 79.1\% & 63.3\% & 46.8\% & 64.1\% & 74.2\% & 61.0\% & 43.2\% & 60.5\% & 65.2\% & 53.6\% & 39.8\% & 53.7\% \\
            TextDiff-200 \cite{liu2023textdiff} & 80.8\% & 66.5\% & 48.7\% & 66.4\% & 77.7\% & 62.5\% & 44.6\% & 62.7\% & 64.8\% & 55.4\% & 39.9\% & 54.2\% \\
            RTSRN-3 \cite{zhang2023pixel} & 80.4\% & 66.1\% & 49.1\% & 66.2\% & 77.1\% & 63.3\% & 46.5\% & 63.2\% & 67.0\% & 59.2\% & 42.6\% & 57.0\% \\
            TCDM \cite{noguchi2024scene} & 81.3\% & 65.1\% & 50.1\% & 66.5\% & 77.6\% & 62.9\% & 45.9\% & 63.1\% & 67.3\% & 57.3\% & 42.7\% & 56.5\% \\
            RGDiffSR \cite{zhou2023recognition} & 81.1\% & 65.4\% & 49.1\% & 66.2\% & 78.6\% & 62.1\% & 45.4\% & 63.1\% & 67.6\% & 56.5\% & 42.7\% & 56.4\% \\
            \midrule
            TATT \cite{ma2022text} & 78.9\% & 63.4\% & 45.4\% & 63.6\% & 72.5\% & 60.2\% & 43.1\% & 59.5\% & 62.6\% & 53.4\% & 39.8\% & 52.6\% \\
            TATT \cite{ma2022text} w/ Ours & \textbf{81.5\%} & \textbf{68.4\%} & \textbf{51.5\%} & \textbf{68.1\%} & \textbf{76.4\%} & \textbf{65.7\%} & \textbf{49.3\%} & \textbf{64.6\%} & \textbf{66.5\%} & \textbf{60.8\%} & \textbf{45.8\%} & \textbf{58.3\%} \\
            \multicolumn{1}{c|}{$\Delta$} & \textbf{\color{red}+2.6\%} & \textbf{\color{red}+5.0\%} & \textbf{\color{red}+6.1\%} & \textbf{\color{red}+4.5\%} & \textbf{\color{red}+3.9\%} & \textbf{\color{red}+5.5\%} & \textbf{\color{red}+6.2\%} & \textbf{\color{red}+5.1\%} & \textbf{\color{red}+3.9\%} & \textbf{\color{red}+7.4\%} & \textbf{\color{red}+6.0\%} & \textbf{\color{red}+5.7\%} \\
            \midrule
            LEMMA \cite{guo2023towards} & 81.1\% & 66.3\% & 47.4\% & 66.0\% & 77.7\% & 64.4\% & 44.6\% & 63.2\% & 67.1\% & 58.8\% & 40.6\% & 56.3\% \\
            LEMMA \cite{guo2023towards} w/ Ours & \textbf{81.9\%} & \textbf{68.3\%} & \textbf{50.7\%} & \textbf{67.9\%} & \textbf{78.6\%} & \textbf{65.6\%} & \textbf{47.9\%} & \textbf{65.0\%} & \textbf{68.1\%} & \textbf{59.8\%} & \textbf{44.4\%} & \textbf{58.1\%} \\
            \multicolumn{1}{c|}{$\Delta$} & \textbf{\color{red}+0.8\%} & \textbf{\color{red}+2.0\%} & \textbf{\color{red}+3.3\%} & \textbf{\color{red}+1.9\%} & \textbf{\color{red}+0.9\%} & \textbf{\color{red}+1.2\%} & \textbf{\color{red}+3.3\%} & \textbf{\color{red}+1.8\%} & \textbf{\color{red}+1.0\%} & \textbf{\color{red}+1.0\%} & \textbf{\color{red}+3.8\%} & \textbf{\color{red}+1.8\%} \\
            \midrule
            HR & 94.2\% & 87.7\% & 76.2\% & 86.6\% & 91.2\% & 85.3\% & 74.2\% & 84.1\% & 76.4\% & 75.1\% & 64.6\% & 72.4\% \\
            \bottomrule
        \end{tabular}
        \caption{Recognition accuracies of various mainstream STISR methods across the three subsets of TextZoom \cite{wang2020scene}. Overall refers to overall accuracy. Best scores are bold.}
        \label{tab:recognition_accuracy}
        \vspace{-0.2cm}
\end{table*}

\subsection{Implementation Details}
We built our proposed method on TATT \cite{ma2022text} and LEMMA \cite{guo2023towards} respectively.
We use the same hyperparameters used in the formal paper in the official implementation of the model.
All of our models are run on PyTorch version 1.13.1.
All experiments are performed on a single NVIDIA RTX A6000 GPU.
The experimental settings of the two works used as baselines, TATT \cite{ma2022text} and LEMMA \cite{guo2023towards}, are completely identical, and the implementation details (batch size, running rate, optimizer, embedding dimension, etc.) will be covered in the supplementary materials.
In TATT \cite{ma2022text}, the pre-trained recognizer is changed from CRNN \cite{shi2016end} to PARSeq \cite{bautista2022scene}.
The loss function for distillation in both models, $\alpha$ is set to 0.5, $\beta$ is set to 0.7, and $\tau$ is set to 3.

\subsection{Ablation Studies}
We examine the impact and efficiency of NCAP and the effectiveness of each module.
And we compare it to methods like label smoothing \cite{he2015delving}, which have been used to address the overconfidence issue.
All evaluations are conducted on TextZoom \cite{wang2020scene}, the STISR dataset, and the text recognizer used for evaluation is CRNN \cite{shi2016end}.

\textbf{Impact on NCAP.}
Our performance evaluations focus on two aspects to assess the influence of NCAP: visualizing results that overcome prior knowledge instability, and analyzing the correlation between prior knowledge error rates and STISR error rates. First, Figure 1 shows that NCAP ensures correct STISR outcomes despite incorrect prior knowledge. Second, Table 2 illustrates experiments on NCAP's dependency. In TATT \cite{ma2022text}, trained only with the teacher's soft label, the proposed method lowers the error rate of prior knowledge and reduces its correlation with STISR errors. Similarly, in LEMMA \cite{guo2023towards}, even though the error rate of prior knowledge increases, the Pearson correlation between prior knowledge errors and STISR errors decreases. Thus, our method no longer relies on unstable text categorical information.

\textbf{Efficiency of NCAP.}
We compare the increase in the number of parameters required to process penultimate layer representations instead of text logits.
Table 3 specifies the overall computational complexity and trainable parameters. When NCAP is introduced to TATT \cite{ma2022text} and LEMMA \cite{guo2023towards}, utilized as baselines, the average increase in overhead is only $0.3\%$.

\textbf{Effectiveness of Each Module.}
We evaluated the effectiveness of each module by sequentially adding them.
As shown in Table 4, the best performance was achieved when all proposed methods were applied. Implementing the linear combination loss of hard and soft labels, an effective method for mitigating the overconfidence issue in the text knowledge modality, resulted in an average accuracy improvement by approximately 3.4\% for TATT \cite{ma2022text} and by about 1.2\% for LEMMA \cite{guo2023towards}. With the additional application of NCAP, there was an improvement by around 3.9\% for TATT \cite{ma2022text} and by about 1.5\% for LEMMA \cite{guo2023towards}.
Finally, by replacing the existing MLP projector with a $\textrm{Conv}_{1 \times 1}$, we achieved an average performance improvement by 5.0\% compared to TATT \cite{ma2022text} and 1.8\% compared to LEMMA \cite{guo2023towards}.

\begin{figure*}[t]
	\centerline{\includegraphics[width=0.9\linewidth]{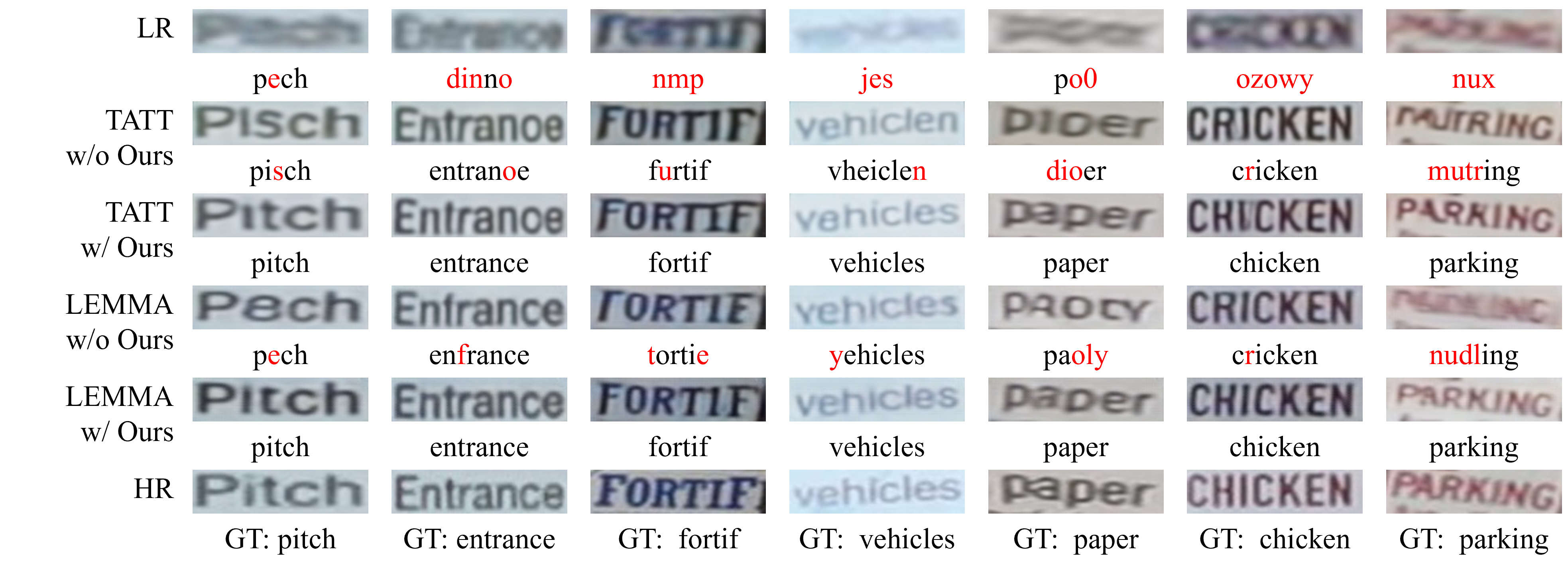}}
	\caption{Visualization of SR images and recognition result on TextZoom \cite{wang2020scene} by CRNN \cite{shi2016end}. \textcolor{red}{Red} indicates wrong recognition results.}
	\vspace{-0.2cm}
\end{figure*}

\begin{table*}[ht]
    \centering\scriptsize
    \begin{tabular}{l|cccc|cccc}
        \toprule
        \multirow{2}{*}{\textbf{Method}} & \multicolumn{4}{c|}{\textbf{Light Degradation}} & \multicolumn{4}{c}{\textbf{Severe Degradation}} \\
        \cmidrule(lr){2-5} \cmidrule(lr){6-9}
        & \textbf{IIIT5K \cite{mishra2012scene}} & \textbf{IC15 \cite{karatzas2015icdar}} & \textbf{SVT \cite{wang2011end}} & \textbf{SVTP \cite{phan2013recognizing}} & \textbf{IIIT5K \cite{mishra2012scene}} & \textbf{IC15 \cite{karatzas2015icdar}} & \textbf{SVT \cite{wang2011end}} & \textbf{SVTP \cite{phan2013recognizing}} \\
        \midrule
        BICUBIC & 34.3\% & 36.6\% & 5.7\% & 62.5\% & 1.3\% & 0.1\% & 0.0\% & 11.1\% \\
        \midrule
        TATT \cite{ma2022text} & 55.1\% & 50.4\% & 44.7\% & 79.3\% & 16.3\% & 11.0\% & 0.0\% & 30.9\% \\
        TATT \cite{ma2022text} w/ Ours & \textbf{60.7\%} & \textbf{58.9\%} & \textbf{68.3\%} & \textbf{80.5\%} & \textbf{22.1\%} & \textbf{13.4\%} & \textbf{3.0\%} & \textbf{32.1\%} \\
        \multicolumn{1}{c|}{$\Delta$} & \textbf{\color{red}+5.6\%} & \textbf{\color{red}+8.6\%} & \textbf{\color{red}+23.7\%} & \textbf{\color{red}+1.1\%} & \textbf{\color{red}+5.8\%} & \textbf{\color{red}+2.4\%} & \textbf{\color{red}+3.0\%} & \textbf{\color{red}+1.1\%} \\
        \midrule
        LEMMA \cite{guo2023towards} & 23.0\% & 21.4\% & 23.5\% & 36.8\% & 9.0\% & 3.4\% & 0.0\% & 18.0\% \\
        LEMMA \cite{guo2023towards} w/ Ours & \textbf{63.0\%} & \textbf{62.6\%} & \textbf{47.1\%} & \textbf{79.2\%} & \textbf{20.4\%} & \textbf{12.8\%} & \textbf{0.3\%} & \textbf{34.0\%} \\
        \multicolumn{1}{c|}{$\Delta$} & \textbf{\color{red}+40.0\%} & \textbf{\color{red}+41.2\%} & \textbf{\color{red}+23.6\%} & \textbf{\color{red}+42.4\%} & \textbf{\color{red}+11.4\%} & \textbf{\color{red}+9.3\%} & \textbf{\color{red}+0.3\%} & \textbf{\color{red}+15.9\%} \\
        \bottomrule
    \end{tabular}
    \caption{Recognition accuracies on scene text recognition datasets with manual degradation applied. We divide the degradation parameters into two categories: light degradation and severe degradation. Manual degradation includes Gaussian blur and Gaussian noise.}
    \label{tab:degradation_results}
    \vspace{-0.2cm}
\end{table*}

\begin{table}[ht]
    \centering\scriptsize
        \begin{tabular}{l|c|c}
            \toprule
            \textbf{Method} & \textbf{PSNR} & \textbf{SSIM} \\
            \midrule
            TATT \cite{ma2022text} & 21.52 & 0.7930 \\
            TATT \cite{ma2022text} w/ Ours & 21.53 & 0.7925 \\
            \midrule
            LEMMA \cite{guo2023towards} & 20.90 & 0.7792 \\
            LEMMA \cite{guo2023towards} w/ Ours & 20.55 & 0.7707 \\
            \bottomrule
        \end{tabular}
        \caption{Comparison of PSNR and SSIM with and without applying the proposed method.}
        \label{tab:psnr_ssim_comparison}
        \vspace{-0.2cm}
\end{table}

\textbf{Different Choices on Overcoming the Domain Gap.}
We aim to analyze the effects of methods considered effective in overcoming the overconfidence issue, including no fine-tuning loss and label smoothing  \cite{he2015delving}.
Table 1 displays the performance change when each technique is added to the baseline. The most conventional method involves using a fixed text recognizer. However, as outlined in TPGSR \cite{ma2023text}, the text prior generated from a fixed text recognizer does not induce overconfidence, but its performance is inherently low as it struggles to bridge the domain gap between LR and HR images. In label smoothing \cite{he2015delving}, the ground-truth hard text label is converted into a soft label by uniformly distributing probabilities across other classes without using a teacher network. However, it fails to address overconfidence and results in poor performance.

\subsection{Comparison with State-of-the-Arts}
We initially assess TextZoom \cite{wang2020scene}, a widely used STISR dataset, by evaluating the text recognition accuracy of restored images using pre-trained text recognizers CRNN \cite{shi2016end}, ASTER \cite{shi2018aster}, and MORAN \cite{luo2019moran}. Additionally, we evaluate the generalization capabilities using STR datasets that sampled LR images.

\textbf{Results on TextZoom.}
We focused on evaluating recognition accuracy, and the proposed method showed consistent improvements across all three pre-trained recognizers. As shown in Table 5, ASTER \cite{shi2018aster}, MORAN \cite{luo2019moran}, and CRNN \cite{shi2016end} with TATT \cite{ma2022text} improved by 4.5\%, 5.1\%, and 5.7\%, respectively, and LEMMA \cite{guo2023towards} improved by 1.9\%, 1.8\%, and 1.8\%. The visualization of SR images and recognition results is shown in Figure 5.

\textbf{Results on STR datasets.}
We evaluate our method’s effectiveness in transforming low-resolution images for text recognition using a scene text recognition dataset and assess generalization with TextZoom \cite{wang2020scene}. We sample images smaller than $16\times64$ and apply manual degradation, including Gaussian blur and Gaussian noise. Since LEMMA \cite{guo2023towards} uses random hyperparameters, we standardized them by categorizing degradation into light and severe. Detailed setup and hyperparameters are in the supplementary materials.
Table 6 shows our method improves TATT \cite{ma2022text} by 13.7\% and LEMMA \cite{guo2023towards} by 23.0\% across both categories.

\section{Discussion}
By applying our proposed method to existing models like TATT \cite{ma2022text} and LEMMA \cite{guo2023towards}, we achieved significant accuracy improvements. However, as shown in Table 7, traditional STISR evaluation metrics such as PSNR and SSIM remained similar or slightly decreased. This is because our approach, which aims to improve readability and accuracy by further reducing noise, can lower visual evaluation scores due to the inherent noise in ground-truth images.
While LEMMA \cite{guo2023towards} also discusses the trade-off between accuracy and visual metrics, the produced SR images perform better than visual metrics.
Visualization and detailed analysis are provided in the supplementary materials.

\section{Conclusion}
In this paper, we identify two key issues with the use of text priors in pre-trained text recognizers: the explicit text prior can cause inconsistencies in STISR tasks, and the method of bridging the domain gap between LR and HR images leads to overconfidence and poor performance. To address this, we propose Non-CAtegorical Prior (NCAP), which replaces unstable text priors with category-free representations. Through theoretical and empirical analysis of prior knowledge, we show that a smoother categorical distribution produces better results. Our method surpasses state-of-the-art approaches, improving performance across all metrics and generalizing well to the STR dataset. It is compatible with all STISR networks using explicit priors.

{\small
\nocite{*}
\bibliographystyle{ieee_fullname}
\bibliography{Untitled}
}

\end{document}